\documentclass[10pt,final,conference,letterpaper,romanappendices,twocolumn]{IEEEtran}
\IEEEoverridecommandlockouts
\usepackage{cite}
\usepackage{graphicx}
\usepackage{textcomp}
\usepackage{xcolor}
\usepackage{epstopdf}
\usepackage{epsfig}
\usepackage{amsthm}
\usepackage{amsmath,amssymb,amsfonts,amstext,amsbsy,amsopn,dsfont}
\usepackage{enumitem}
\usepackage{cases}
\usepackage{sublabel}

\newtheorem{theorem}{\textbf{Theorem}}

\newtheorem{definition}{\textbf{Definition}}

\newcommand{\defn}{\triangleq}

\begin{document}

\title{Spatio-Temporal Federated Learning for Massive Wireless Edge Networks\\
\thanks{The work of C.-H. Liu was supported in part by the National Science Foundation under Award CNS-2006453 and in part by Mississippi State University under Grant ORED 253551-060702. The work of L. Wei was supported in part by the National Science Foundation ($\#$2150486 and $\#$2006612). The work of K.-T. Feng was supported in part by the Ministry of Science and Technology (MoST) under Grants 110-2221-E-A49-041-MY3, 110-2224-E-A49-001, Higher Education Sprout Project of the National Yang Ming Chiao Tung University and Ministry of Education (MoE), Taiwan.}
}

\author{\IEEEauthorblockN{Chun-Hung Liu$^{\dagger}$, Kai-Ten Feng$^{\ddagger}$, Lu Wei$^{*}$, and Yu Luo$^{\dagger}$}
\IEEEauthorblockA{Department of Electrical \& Computer Engineering, Mississippi State University, MS, USA$^{\dagger}$ \\ Department of Electrical \& Computer Engineering, National Yang Ming Chiao Tung University, Hsinchu, Taiwan$^{\ddagger}$ \\ Department of Computer Science, Texas Tech University, TX, USA$^{*}$ \\
e-mail: \{chliu, yu.luo\}@ece.msstate.edu$^{\dagger}$; ktfeng@nycu.edu.tw$^{\ddagger}$; luwei@ttu.edu$^{*}$}
}

\maketitle

\begin{abstract}
This paper presents a novel approach to conduct highly efficient federated learning (FL) over a massive wireless edge network, where an edge server and numerous mobile devices (clients) jointly learn a global model without transporting the huge amount of data collected by the mobile devices to the edge server. The proposed FL approach is referred to as spatio-temporal FL (STFL), which jointly exploits the spatial and temporal correlations between the learning updates from different mobile devices scheduled to join STFL in various training epochs. The STFL model not only represents the realistic intermittent learning behavior from the edge server to the mobile devices due to data delivery outage, but also features a mechanism of compensating loss learning updates in order to mitigate the impacts of intermittent learning. An analytical framework of STFL is proposed and employed to study the learning capability of STFL via its convergence performance. In particular, we have assessed the impact of data delivery outage, intermittent learning mitigation, and statistical heterogeneity of datasets on the convergence performance of STFL. The results provide crucial insights into the design and analysis of STFL-based wireless networks.
\end{abstract}


\section{Introduction}
Conventional machine learning techniques typically work in a centralized manner, where all clients transport the data to a central server to execute learning. Although such centralized learning techniques achieve the desired learning performance while exploiting a large amount of data from clients, they may suffer risk of transporting privacy-sensitive data. For example, centralized machine learning may not be appropriate in medical data analysis since the data of patients can not be shared or transported due to privacy and security concerns. In addition, implementing centralized machine learning over wireless networks induces other practical challenges, such as large transmission latency, heavy traffic load, and large resource consumption. For example, in addition to suffering a long latency, mobile devices may consume substantial amount of power to upload their datasets to the radio access network. This is because the backbone networks to the cloud are not optimally designed for delay-sensitive services~\cite{SCHULZ17,KC15wc}. Moreover, the large amount of wireless data transportation consumes considerable network resources, thereby leading to significant networking latency. To address these issues, federated learning (FL) over wireless communication becomes a promising candidate in both academia and industry.

The mechanism of FL over a wireless network is to enable all wireless clients to train a local model using their own dataset before uploading the locally trained model to a server for global model aggregation. The server then sends the aggregated model back to all the wireless clients for the next training round. As the training process between the wireless clients and the server proceeds, the locally trained models are expected to converge to a global model. The study of FL over wireless networks is still in its infancy, yet it can be categorized into three main directions: over-the-air digital and analog data aggregation, communication and computation efficiency, privacy and security~\cite{YQLHJCXG19,JRHWTH19,KYTJYS2020,JRENE2019,SWE2019,DG2019,MMADG20-1,MMADG20-2,GZYDDG21,CLGLPKV21}. References~\cite{MMADG20-1} and~\cite{MMADG20-2}, for example, studied how to achieve FL through digital and analog signal combining techniques in wireless channels. The works~\cite{SWE2019,JRENE2019} focused on the computation off-loading in FL, whereas the problems regarding the reduction of edge computing and communication efficiency were addressed in~\cite{HBMEMDR16,JKHBMFXY16}. 

\begin{figure*}[!t]
	\centering
	\includegraphics[width=0.95\textwidth, height=1.75in]{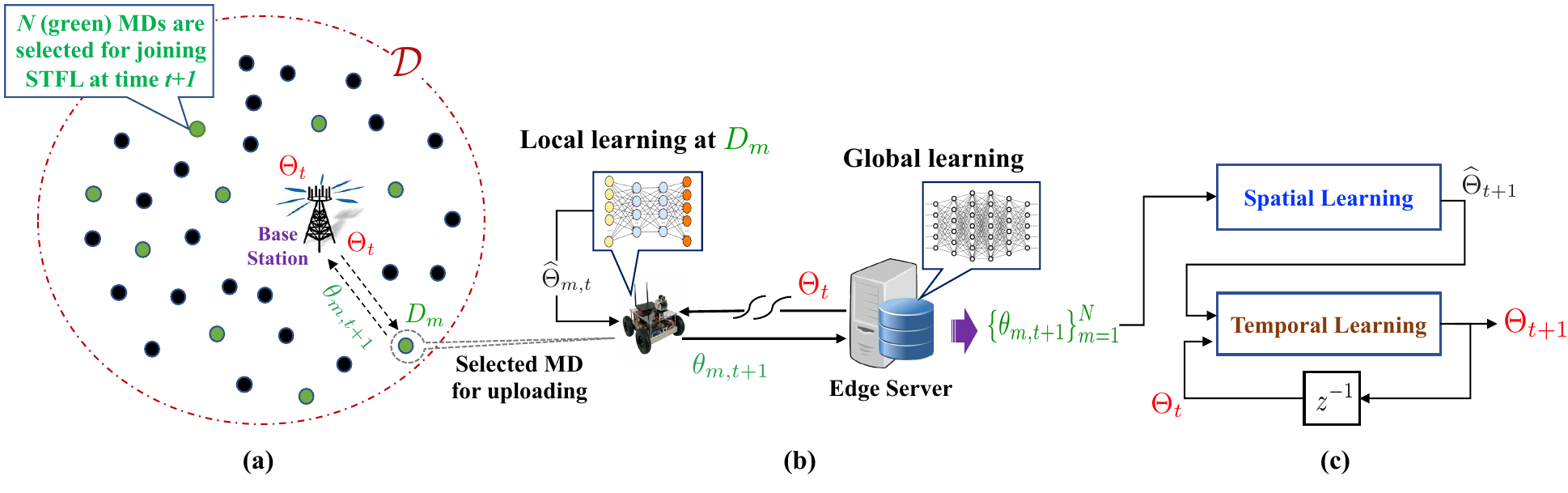}
	\vspace{-6pt}
	\caption{\small (a) A massive wireless edge network in which a base station randomly selects $N$ MDs from set $\mathcal{D}$ to join STFL in each time slot. (b) The model of STFL between the edge server and $D_m$ selected to upload its local learning outcome $\theta_{m,t+1}$ in time slot $t+1$. (c) The block diagram of the spatio-temporal learning model at the edge server.}
	\label{Fig:ModelLSFEL}
	\vspace{-16pt}
\end{figure*}

Most existing works in the literature including the aforementioned ones have not studied how to efficiently conduct FL over a massive wireless network, such as an internet-of-thing (IoT) network, where numerous IoT (mobile) devices jointly conduct FL under the circumstance of limited radio and/or networking resources. Therefore, there is an urgent need to devise a new large-scale FL methodology to efficiently exploit the huge amount of data stored in massive wireless networks. Our main contribution in this paper is to propose a novel FL model for a massive wireless edge network with a large number of mobile devices (clients), referred to as spatio-temporal FL (STFL), which aggregates different locally trained models into a global learning model by exploiting their spatial and temporal correlations. In addition, the proposed STFL model considers a realistic intermittent learning scenario from an edge server to mobile devices due to data delivery outage, and employs a mechanism of mitigating the impact of intermittent learning. We also propose a definition of learning capability to devise a framework of analyzing the convergence performance of STFL. We derive a sufficient condition that ensures STFL to achieve its learning capability, which further leads to the corresponding transient analysis of STFL. Our analytical findings quantify the fundamental relationships between data delivery outage, effect of mitigating intermittent learning, and statistical heterogeneity of data in different datasets. The results provide profound insights into the improvement of the convergence rate of STFL.

\section{Model of Spatio-Temporal Federated Learning}\label{Sec:SystemModel}
We consider a massive wireless edge network consisting of a base station (BS) and a large set of mobile devices (MDs) denoted by $\mathcal{D}$. The BS is connected to an edge server with a learning facility and helps the information exchange between the edge server and the MDs in set $\mathcal{D}$. Each MD possesses a dataset to perform a local learning algorithm. The edge server and all the MDs in the edge network aim to learn a global model (vector) by utilizing the data stored in all the datasets of the MDs. Due to privacy and other practical concerns, none of the datasets of the MDs can be transported to or accessed by any other devices in the edge network. Since set $\mathcal{D}$ is fairly large and the radio resource is limited, current prevailing FL models in the literature may not be directly adopted in such a scenario. In order to efficiently exploit the large amount of data stored in $\mathcal{D}$, we propose a new model of FL between the edge server and set $\mathcal{D}$ as follows. In each training epoch of the learning process, the BS randomly selects $N$ MDs from $\mathcal{D}$ to join the learning process, which is merely a small portion of the MDs in $\mathcal{D}$. Each of the selected MDs utilizes its own dataset to train the local learning model before uploading its training outcome to the edge server through the BS. The edge server then aggregates all the received local models into a global model and sends it back to \textit{all} the MDs to complete a training epoch. 



To ensure the learning process will efficiently exploit the data from the $N$ selected datasets, we propose a novel FL model as follows. Suppose MD $m$ is selected to upload its local learning model at time $t$ and its dataset can be expressed as
\begin{align}\label{Eqn:DataSet}
\mathcal{S}_m\defn \{ & S_{m,i}\in\mathbb{R}^d\times\mathbb{R}: S_{m,i}=(x_{m,i},y_{m,i}), \nonumber\\
& x_{m,i}\in\mathbb{R}^d, y_{m,i}\in\mathbb{R}\},
\end{align}
where $S_{m,i}$ denotes data point $i$ in dataset $\mathcal{S}_m$, $x_{m,i}$ is a $d$-dimensional $(d$-D) data vector with $d$ feature elements, and $y_{m,i}$ is the labeled (scalar) output corresponding to $x_{m,i}$. At (training epoch) time $t$, MD $D_m$ receives the global model $\Theta_t\in\mathbb{R}^d$ sent by the BS and uses this $\Theta_t$ to update its local learning model $\theta_{m,t+1}$ via the following algorithm:
\begin{align}\label{Eqn:LocalLearnModel}
\hspace{-8pt}\begin{cases}
\Theta_{m,t} &= \gamma_{m,t}\Theta_t+(1-\gamma_{m,t}) \widehat{\Theta}_{m,t}\\
\mathbf{g}_m(\Theta)&= \frac{1}{|\mathcal{S}_m|}\sum_{S_{m,i}\in\mathcal{S}_m}\nabla_{\Theta}\ell(S_{m,i},\Theta)\\
\theta_{m,t+1} &= \Theta_{m,t} - \alpha_m \mathbf{g}_m\left(\Theta_{m.t}\right)
\end{cases}.
\end{align}
Here, $\gamma_{m,t}\in\{0,1\}$ is a Bernoulli random variable that characterizes the data delivery outage from the edge server to $D_m$\footnote{The data delivery outage from the server to an MD could be caused, among others, by wireless channel fading, network congestion, and queuing delay.}, $\widehat{\Theta}_{m,t}$ is the estimate of $\Theta_t$ once $\Theta_t$ is lost, $\alpha_m>0$ denotes the \textit{spatial } learning rate for MD $m$, $|\mathcal{S}_m|$ denotes the number of the data points in $\mathcal{S}_m$, $\ell(\cdot,\cdot): \mathbb{R}^d\times\mathbb{R}\rightarrow \mathbb{R}_+$ is a differentiable \textit{loss} function\footnote{The loss function $\ell(\cdot,\cdot)$ can be generally designed to represent various learning models adopting a stochastic gradient decent method, e.g., support vector machines, neural networks, and linear regression.}, and $\nabla_{\Theta}\ell(\cdot,\Theta)$ represents the gradient of $\ell(\cdot,\Theta)$ with respect to argument $\Theta$. The function $\widehat{\Theta}_{m,t}$ is to compensate $\Theta_t$ when MD $D_m$ does not receive it at time $t$ and we will demonstrate how to determine $\widehat{\Theta}_{m,t}$ in Section~\ref{Sec:ConvAnaSTFL}. The learning model between the edge server and MD $D_m$ in a massive wireless edge network is illustrated in Fig.~\ref{Fig:ModelLSFEL}(a) and (b). 

At time $t+1$, the $N$ MDs selected by the BS upload their locally trained models to the edge server through the BS. The edge server then adopts the following algorithm to find $\Theta_{t+1}$,
\begin{align}\label{Eqn:GlobalLearnModel}
\begin{cases}
\widehat{\Theta}_{t+1} &= \sum_{m=1}^{N}\frac{ |\mathcal{S}_{m,t}|}{|\mathcal{S}_t|}\theta_{m,t+1}\\
\Theta_{t+1} &= \Theta_t-\beta_t  \left(\Theta_t-\widehat{\Theta}_{t+1}\right)  
\end{cases},
\end{align}
where $\beta_t\in(0,1)$ is known as the \textit{temporal} learning rate, $\mathcal{S}_{m,t}$ denotes the dataset of an MD $D_m$ selected at time $t$, and $|\mathcal{S}_t|\defn \sum_{m=1}^{N}|\mathcal{S}_{m,t}|$. The algorithm for $\widehat{\Theta}_{t+1}$ is the spatial learning over the $N$ datasets $\{\mathcal{S}_{m,t}\}_{m=1}^N$, whereas the recursive algorithm for $\Theta_{t+1}$ conducts the temporal learning over all $\overline{\Theta}_t$ before time $t+1$. The algorithm in~\eqref{Eqn:GlobalLearnModel} is motivated by the idea of the moving average among the $t$ global learning models found prior to time $t+1$
because $\Theta_t$ in~\eqref{Eqn:GlobalLearnModel} reduces to the moving average of the set $\{\Theta_1,\ldots,\Theta_t\}$ if $\beta_t=1/(t+1)$. For the simplicity of modeling the learning behavior at the edge server, we assume that the data delivery outage from an MD to the edge server in~\eqref{Eqn:GlobalLearnModel} does not impact the learning process of $\Theta_t$\footnote{This is a reasonable assumption as it is very unlikely that many of the $N$ MDs fail to transmit their data to the server at the same time, where the temporal learning also mitigates the effects of data delivery outage.}. \textit{The global learning algorithm in~\eqref{Eqn:GlobalLearnModel} can exploit the spatial (local) learning outcomes across different times and datasets distributed over the edge network} since the global models found at different times are likely based on different local datasets. As a result, the local learning algorithm in~\eqref{Eqn:LocalLearnModel} and the global learning algorithm in~\eqref{Eqn:GlobalLearnModel} are referred to as STFL in this paper. Thus, the proposed STFL model is able to achieve the spatial average over the set $\mathcal{D}$ as $t\rightarrow\infty$ since $\mathcal{D}$ is large according to the mean-field theory~\cite{ABJFPY13}. Note that the proposed STFL model reduces to the general FL model in the literature whenever $\gamma_{m,t}=\beta_t=1$ for all $t\in\mathbb{N}$. This manifests the fact that a general FL model cannot handle data delivery outage in the presence of a huge amount of data widely distributed over a massive edge network. The block diagram of the STFL algorithm is shown in Fig.~\ref{Fig:ModelLSFEL}(c). In the following section, we will study the convergence performance of STFL.

\section{Convergence Analysis of STFL}\label{Sec:ConvAnaSTFL}
In this section, we investigate the convergent performance of the proposed STFL so as to understand how data delivery outages impact federated learning. Let $\Theta_*\in \mathbb{R}^d$ be the \textit{target} global model that all the MDs would like to learn, that is, all the local models in~\eqref{Eqn:LocalLearnModel} are expected to converge to $\Theta_*$ as time goes to infinity. We will first analyze the steady-state behavior of STFL to understand when the proposed STFL algorithm is able to accurately learn the global model. We will then focus on the transient analysis of STFL in understanding the transient behavior and convergence rate.

\subsection{Analysis of the Learning Capability of STFL}
To evaluate if STFL is able to learn in a long-time sense, we propose the notion of ``learning capability'' for the proposed STFL, as defined in the following.
\begin{definition} [Learning capability of STFL]\label{Defn:LocalFELCap}
Let $\epsilon_{m,t}\defn \mathbb{E}\left[\|\theta_{m,t}-\Theta_*\|^2\right]$ be the learning error of the local model $\theta_{m,t}$ of MD $\mathcal{D}_m$ at time $t$. The proposed STFL is said to possess the learning capability if $\epsilon_{m,t}$ converges to zero as $t$ goes to infinity for all $D_m\in\mathcal{D}$. 
\end{definition}
\noindent Definition~\ref{Defn:LocalFELCap} manifests the fact that the proposed STFL possesses a learning capability if the learning error of all the local models is sufficiently small after a long training time. According to this definition, we can find the condition that enables STFL to possess a learning capability, as shown in the following theorem.
\begin{theorem}\label{Thm:ConvSTFL}
If each MD $D_m\in\mathcal{D}$ is able to estimate its missed global model such that $\mathbb{E}[\|\Theta_t-\widehat{\Theta}_{m,t}\|^2]\leq \delta_m\epsilon_{m,t}$ holds for a small $\delta_m>0$ and all $t\in\mathbb{N}$, then the STFL algorithm proposed in~\eqref{Eqn:LocalLearnModel} and~\eqref{Eqn:GlobalLearnModel} possesses the learning capability if the following inequality holds
\begin{align}\label{Eqn:ConvergeCond}
\max_{m:D_m\in\mathcal{D}} \left\{\sqrt{(1+q_m\delta_m)} \sigma_m(\Theta)\right\}<1,
\end{align}
where $q_m=\mathbb{P}[\gamma_{m,t}=0]$ is the outage probability of the data delivery from the edge server to $D_m$ and $\sigma_m(\Theta)$ is the spectral radius of matrix $\mathbf{I}_d-\alpha_m\nabla_{\Theta}\mathbf{g}_m(\Theta)$, and $\mathbf{I}_d$ denotes a $d\times d$ identity matrix.
\end{theorem}
\begin{IEEEproof}
See Appendix~\ref{App:ProofConvSTFL}.
\end{IEEEproof}

The inequality in Theorem~\ref{Thm:ConvSTFL} reveals some important implications, which are worth pointing out in the following. First of all, the result characterizes how the data delivery outage and the accuracy of estimating the missed global model at each MD impact the learning capability of STFL. The data delivery outage brings a negative impact on the learning capability because it increases the right hand side of the inequality in~\eqref{Eqn:ConvergeCond}, which makes \eqref{Eqn:ConvergeCond} more difficult to hold. Nonetheless, such a negative impact can be significantly mitigated if all $\delta_m$'s are very small, which is accomplished whenever each MD is able to accurately estimate and compensate its lost global model. We will propose a stochastic approximation algorithm for each MD to estimate the global model and numerically evaluate its estimation performance in the following section. In addition to accurately compensating the lost global model, another effective method that helps STFL equip with the learning capability is to find an optimal $\alpha_m$ that minimizes $\sqrt{(1+q_m\delta_m)}\sigma_m(\Theta)$. If $\lambda_{m,i}>0$ denotes one of the eigenvalues of $\nabla_{\Theta}\mathbf{g}_m(\Theta)$, then $\sigma_m(\Theta)=\max_i\{|1-\alpha_m\lambda_{m,i}|\}$ and the following optimization problem for $\alpha_m$ can be formulated,
\begin{align}
&\min_{\alpha_m}\left\{\max_{i\in\{1,\ldots,d\}}\left\{|1-\alpha_m\lambda_{m,i}|\right\}\right\}\\
\text{s.t. }&\frac{\sqrt{(1+q_m\delta_m)}-1}{\lambda^{\max}_m\sqrt{(1+q_m\delta_m)}}< \alpha_m< \frac{\sqrt{(1+q_m\delta_m)}+1}{\lambda^{\max}_m\sqrt{(1+q_m\delta_m)}},\label{Eqn:ConsLearnRate}
\end{align}
where~\eqref{Eqn:ConsLearnRate} is derived from the constraint $\sigma_m(\Theta)=\max_i\{|1-\alpha_m\lambda_{m,i}|\}<1$ and $\lambda^{\max}_m\defn\max\{\lambda_{m,1},\ldots,\lambda_{m,d}\}$. The solution to this optimization problem can be found as
\begin{align}\label{Eqn:OptLearnRate}
\alpha^*_m = \frac{2}{\lambda^{\max}_m+\lambda^{\min}_m}
\end{align}
if the following inequality holds
\begin{align*}
1<\frac{\lambda^{\max}_m}{\lambda^{\min}_m}<\frac{\sqrt{(1+q_m\delta_m)}+1}{\sqrt{1+q_m\delta_m}-1},
\end{align*}
where $\lambda^{\min}_m=\min\{\lambda_{m,1},\ldots,\lambda_{m,d}\}$ and $\frac{\lambda^{\max}_m}{\lambda^{\min}_m}$ is known as the condition number of $\nabla_{\Theta}\mathbf{g}_m(\Theta)$. Apparently, the condition number of $\nabla_{\Theta}\mathbf{g}_m(\Theta)$ has a profound impact on the learning capability of STFL. For example, if $\frac{\lambda^{\max}_m}{\lambda^{\min}_m}$ is small and close to one (i.e., $\lambda_{m,1}\approx\lambda_{m,i}\approx\lambda_{m,d}\approx\lambda_m$), $\alpha^*_m\approx \frac{1}{\lambda_{m}}$, thereby $1-\alpha^*_m\lambda_{m,i}\approx 0$. In such a scenario, STFL will readily possess the learning capability by using the optimal learning rate $\alpha^*_m$ even if it may seriously suffer from the data delivery outage. A large condition number usually leads to small $\alpha_m$ in order to enable the learning capability, thereby reducing the convergence rate of STFL. Another important implication that can be learned from~\eqref{Eqn:ConvergeCond} is how the different distributions of the non-i.i.d. datasets in the edge network affect the learning capability. This is because whether or not ~\eqref{Eqn:ConvergeCond} holds hings on the eigenvalues of $\nabla\mathbf{g}_m(\Theta)$ that are dependent on the distribution of the data point in dataset $\mathcal{S}_m$. In general, the term $\sqrt{(1+q_m\delta_m)}\sigma_m(\Theta)$ in~\eqref{Eqn:ConvergeCond} increases and $\alpha^*_m$ in~\eqref{Eqn:OptLearnRate} accordingly decreases as the number of the non-i.i.d. datasets in the edge network increases. Hence, Theorem~\ref{Thm:ConvSTFL} provides an analytical foundation on how non-i.i.d. datasets influences the learning performance. Nonetheless, we are unable to infer the transient behavior of STFL from Theorem~\ref{Thm:ConvSTFL} alone, the study of which is presented in the following subsection.

\subsection{Analysis of the Transient Performance of STFL}

The transient behavior of STFL can be characterized by how fast the local learning models converge within a fixed number of training epochs. Accordingly, we propose the concept of the learning time constant of STFL specified in the following definition. 
\begin{definition}[Learning Time Constant of STFL]
Suppose STFL possesses the learning capability. The local time constant of MD $D_m$, denoted by $\tau_m$, is defined as 
\begin{align}\label{Eqn:DefnLocalTimeConst}
\tau_m \defn \inf\{T: \epsilon_{m,t+T}\leq e^{-1}\epsilon_{m,t}, t\in\mathbb{N}\}.
\end{align}
The learning time constant of STFL is $\tau\defn \max_{m:D_m\in\mathcal{D}}\{\tau_m\}$.
\end{definition}
The physical meaning of the learning time constant $\tau_m$ is the minimum time period needed to make the learning error $\epsilon_{m,t+\tau_m}$ reduce to $36.79\%$ of $\epsilon_{m,t}$, and it is similar to the meaning of the time constant of a linear dynamic system. The learning time constant of STFL is thus defined as the maximum among all the local time constants. Therefore, the key to improving the convergence rate of STF is to reduce the learning time constant. The time constant defined in~\eqref{Eqn:DefnLocalTimeConst} can be equivalently expressed as
\begin{align}
\tau_m = \inf\left\{T: \ln\left(\frac{\epsilon_{m,t}}{\epsilon_{m,t+T}}\right)\geq 1, t\in\mathbb{N}\right\},\label{Eqn:DefnLocalTimeConst2}
\end{align}
which can be explicitly found as shown in the following theorem.
\begin{theorem}\label{Thm:TimeConstSTFL}
If the STFL possesses the learning capability, the local time constant of MD $D_m$ can be found as
\begin{align}
\tau_m = \frac{-1}{2\ln\left(\sqrt{(1+q_m\delta_m)}\sigma^*_m(\Theta)\right)}, \label{Eqn:LocalTimeConst}
\end{align}
where $\sigma^*_m(\Theta)\defn \max_{i\in\{1,\ldots,d\}}\{|1-\alpha^*_m\lambda_{m,i}|\}$ and $\alpha^*_m$ is given in~\eqref{Eqn:OptLearnRate}. The learning time constant of STFL is thus obtained as
\begin{align}
\tau = -\frac{1}{2}\left\{\max_{m:D_m\in\mathcal{D}}\ln\left(\sqrt{(1+q_m\delta_m)} \sigma^*_m(\Theta)\right)\right\}^{-1}. \label{Eqn:LearnTimeConst}
\end{align}
\end{theorem}
\begin{IEEEproof}
See Appendix~\ref{App:ProofAsyConvSTFL}.
\end{IEEEproof}

The outcomes shown in Theorem~\ref{Thm:TimeConstSTFL} clearly indicate how the convergence rate of STFL is quantitatively affected by the data delivery outage, the effect of compensating the lost global model on the MD side, and the statistical property of the datasets in the edge network. To decrease $\tau_m$, we need to keep $\sqrt{(1+q_m\delta_m)}\sigma^*_m(\Theta)$ as small as possible. As such, both the transient behavior and the learning capability of STFL are dominated by the value of $\sqrt{(1+q_m\delta_m)}\sigma^*_m(\Theta)$. The distributions of the non-i.i.d. datasets have a much more critical impact on the transient behavior of STFL than the data delivery outage and the compensating effect. The reason is that $(1+q_m\delta_m)$ can be reduced to one, whereas $\ln(\sigma^*_m(\Theta))$ can be reduced to a value much smaller than one. To demonstrate this, consider a scenario of no data delivery outage (or the lost global learning model is perfectly compensated), the convergence performance of STFL is dominated by the maximum of $\sigma^*_m(\Theta)$,
\begin{align}
\max_m \{\sigma^*_m(\Theta)\} = \max_m\max_i \left\{\bigg|1-\frac{2\lambda_{m,i}}{\lambda^{\max}_{m}+\lambda^{\min}_{m}}\bigg|\right\},
\end{align} 
which approaches to zero as $\lambda^{\min}_{m}$ increases to $\lambda^{\max}_{m}$, and thereby $-\ln(\sigma^*_m(\Theta))$ can be quite large. Namely, $\tau$ can be considerably reduced as the condition number of $\nabla_{\Theta}\mathbf{g}_m(\Theta)$ that depends on $\mathcal{S}_m$ is close to unity.  This suggests that carefully removing some features of the non-crucial input data points helps to reduce the condition number and accordingly improves the convergence rate of STFL. We will illustrate this observation in the following section.

\section{Numerical Results}
\vspace{-2pt}
In this section, some numerical results are provided to illustrate our analytical findings in the previous section. We consider the wireless edge network in Fig.~\ref{Fig:ModelLSFEL}, where $10000$ MDs are uniformly distributed in set $\mathcal{D}$ and the BS randomly schedules $100$ MDs to join the learning process of STFL in each training epoch. For simplicity, a linear regression problem is tackled in the wireless edge network by using STFL. As a result, the loss function becomes $\ell(S_{m,i},\Theta) = \frac{1}{2}(y_{m,i}-\Theta^Tx_{m,i})^2$ so that we obtain $\nabla_{\Theta}\ell(S_{m,i},\Theta) = (\Theta^Tx_{m,i}-y_{m,i})x_{m,i}$ and $\nabla_{\Theta}\mathbf{g}_m(\Theta)=\frac{1}{|\mathcal{S}_m|}\sum_{S_{m,i}\in\mathcal{S}_m}x_{m,i}x^T_{m,i}\approx \Sigma_m+\mathbb{E}[x_m]\mathbb{E}[x_m]^T$, where $\Sigma_m$ is the covariance matrix of all $x_{m,i}$'s in $\mathcal{S}_m$. Each MD adopts the following algorithm to find $\widehat{\Theta}_{m,t}$ in~\eqref{Eqn:LocalLearnModel},
\begin{align}\label{Eqn:CompMissGlobModel}
\widehat{\Theta}_{m,t} = (1-\omega) \sum_{i=1}^{t} \omega^{t-i}\left[(1-\gamma_{m,i}) \theta_{m,i-1}+\gamma_{m,i}\Theta_i\right],
\end{align}
where $\omega\in(0,1)$. Note that $\omega=0$ corresponds to the case that each MD does not compensate its lost global model. Since it is clear that $\lim_{t\rightarrow\infty}\widehat{\Theta}_{m,t}=\Theta_*$ once $\Theta_t\rightarrow\Theta_*$ as $t\rightarrow\infty$, there must exist a fairly small $\delta_m$ such that $\|\widehat{\Theta}_{m,t}-\Theta_t\|\leq \delta_m \|\theta_{m,t}-\Theta_*\|=\delta_m\epsilon_{m,t}$. In the following simulation, $\omega=0.25$ will be used in \eqref{Eqn:CompMissGlobModel} if MD $D_m$ needs to compensate its lost global model and consider $\delta_m\leq 1$. Moreover, all the datasets in the network are of the same size with $100$ data points, i.e., $|\mathcal{S}_m|=100$ and $x_{m,i}\in\mathbb{R}^3$ for $m=1,2,\ldots,10000$ and $i=1,2,\ldots,100$. The datasets are non-i.i.d. and equally likely share two different $2$-D Gaussian distributions with two distinct means
\begin{align}\label{Eqn:VecMean}
\mathbb{E}[x_{m,i}]\in
\left\{
\left[\begin{array}{c}
1 \\
1 
\end{array}\right],  
\left[\begin{array}{c}
2.2 \\
1.8 
\end{array}\right]
\right\},
\end{align} 
 and covariance matrices
\begin{align}\label{Eqn:CovMat}
\hspace{-10pt}\Sigma_m \in\left\{\left[\begin{array}{cc}
	1 & 1.25  \\
	1.25 & 3 
\end{array}
\right], \left[\begin{array}{cc}
	2 & 1.75 \\
	1.75 & 2 
\end{array} 
\right]
\right\}.
\end{align}
The STFL algorithm aims to learn a $2$-D global model $\Theta_*=\frac{1}{\sqrt{2}}[1\,\, -1]^T$. Accordingly, all the input data points $x_{m,i}$ in $\mathcal{S}_m$ are generated by using~\eqref{Eqn:VecMean} and~\eqref{Eqn:CovMat}, and all the output data points $y_{m,i}$ in $\mathcal{S}_m$ are generated by $y_{m,i}=\Theta^T_*x_{m,i}$. 

\begin{figure}[!t]
	\centering
	\vspace{-4pt}
	\includegraphics[height=1.85in,width=1.0\linewidth]{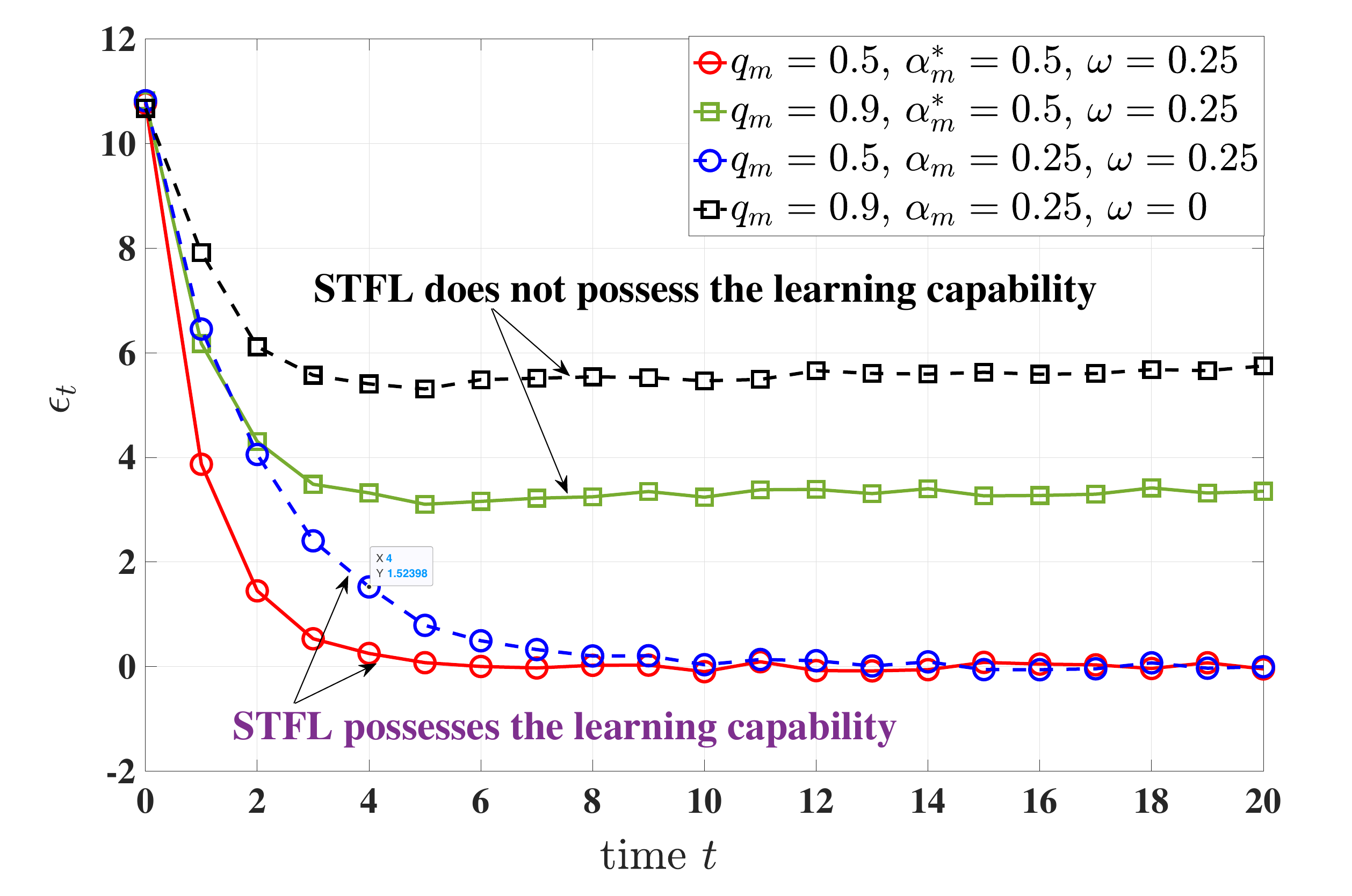}
	\vspace{-18pt}
	\caption{The simulation results of the averaged learning error $\epsilon_t$.}
	\label{Fig:LearningError}
	\vspace{-16pt}
\end{figure}

\begin{figure}[!t]
	\centering
	\vspace{-6pt}
	\includegraphics[height=1.8in,width=1.0\linewidth]{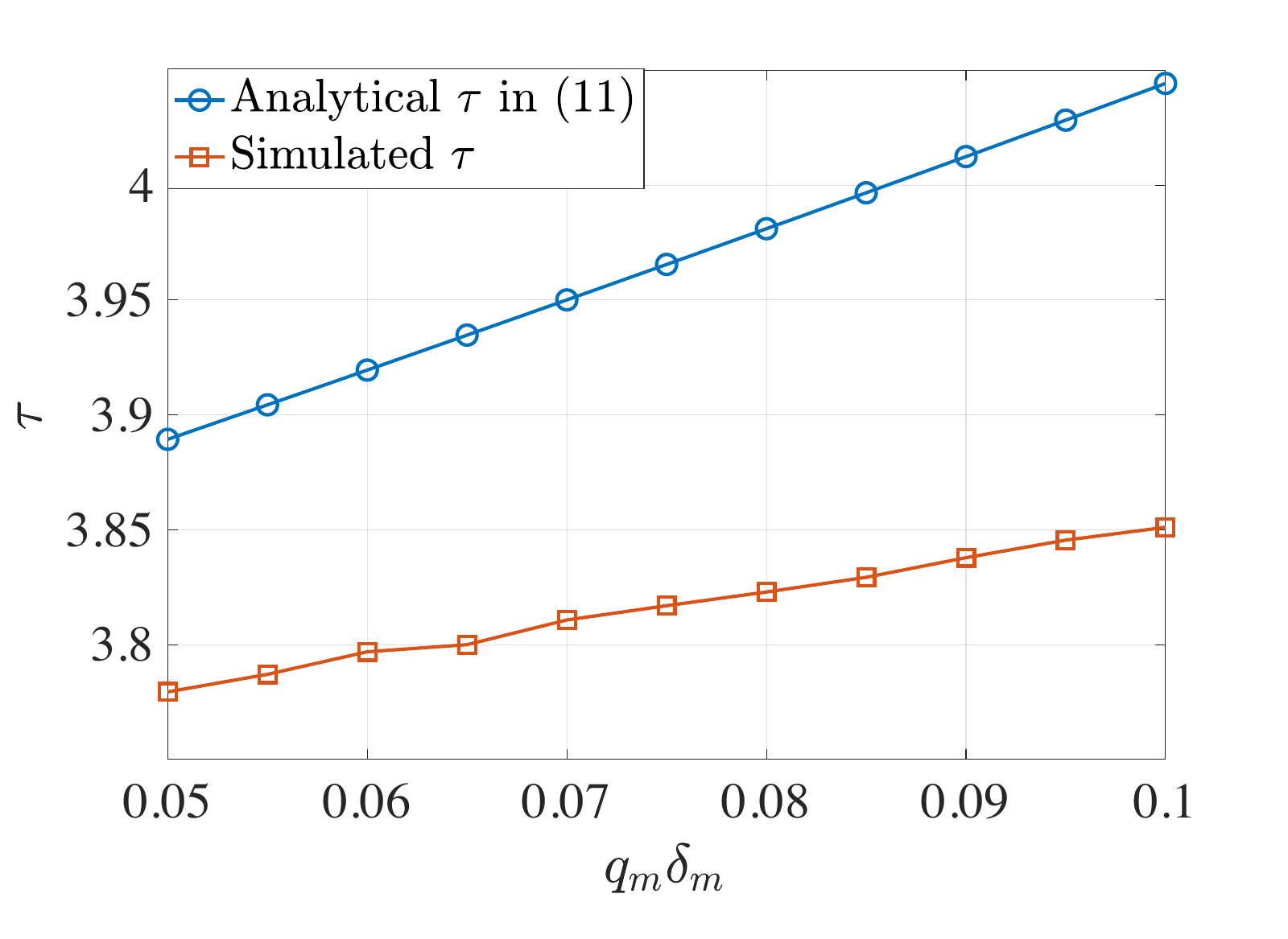}
	\vspace{-22pt}
	\caption{The simulation results of the learning time constant $\tau$. }
	\label{Fig:TimeConst}
	\vspace{-12pt}
\end{figure}

According to~\eqref{Eqn:OptLearnRate} and \eqref{Eqn:CovMat}, we can find $\alpha^*_m\approx 0.5$ for all $D_m\in\mathcal{D}$. We adopt $\beta_t = 1/(t+1)$ in~\eqref{Eqn:GlobalLearnModel}, assume the same $q_m$ for all $D_m\in\mathcal{D}$, and define the averaged learning error as $\epsilon_t = \frac{1}{N}\sum_{m=1}^{N}\epsilon_{m,t}$. The simulation results of $\epsilon_t$ is shown in Fig.~\ref{Fig:LearningError} for four different combinations of $q_m$ and $\alpha_m$. As shown in the figure, all the curves eventually converge, yet two of them (i.e., the curves with squares) do not converge to zero. Namely, the two combinations, $q_m=0.9,\alpha_m=0.25$ and $q_m=0.9,\alpha^*_m=0.5$, do not make STFL possess the learning capability because they do not satisfy \eqref{Eqn:ConvergeCond}. Furthermore, the dash-square curve is the case of no global model compensation so that its $\epsilon_t$ apparently converges to a much higher value than the other three. The other two combinations satisfy~\eqref{Eqn:ConvergeCond} and indeed their curves converge to zero. Also, the circle-solid (red) curve with optimal $\alpha^*_m=0.5$ converges much faster than the circle-dash (blue) one without $\alpha^*_m$, as expected.  Fig.~\ref{Fig:TimeConst} shows the simulation results of how the learning time constant $\tau$ varies with $q_m\delta_m$. As can be seen from the figure, the learning time constant found in~\eqref{Eqn:LearnTimeConst} is slightly larger than the simulated one. This is because we use a (tight) bound when deriving $\tau$. Nonetheless, $\tau$ in~\eqref{Eqn:LearnTimeConst} is still very accurate so that it provides a useful evaluation on how the convergence rate of STFL is affected under different values of $q_m\delta_m$.

\section{Conclusion}
To effectively learn from the data stored in a massive edge network with a limited radio resource, we devised a novel STFL technique, which is able to spatially and temporally conduct the learning process among a large number of MDs. The salient features of STFL lie in two facets of modelling the realistic conditions incurred by massive wireless networking of unreliable communication and limited radio resource. The proposed STFL algorithm employs a mechanism of compensating lost global models at each MD as well as scheduling MDs at the edge server. An analytical framework for characterizing the learning capability of STFL was proposed and employed to find a sufficient condition for STFL convergence in a long-term sense. The learning time constant of STFL was defined and derived in order to evaluate the transient performance of STFL. Our analytical findings shed new light on the fundamental interplay between data delivery outage, compensation of lost data, and distribution of non-i.i.d. datasets. 

\appendix
\numberwithin{equation}{section}
\setcounter{equation}{0}

\subsection{Proof of Theorem~\ref{Thm:ConvSTFL}}\label{App:ProofConvSTFL}
First of all, we adopt the notation $\Delta \mathbf{z}\defn \mathbf{z}-\Theta_*$ for the difference between any vector $\mathbf{z}$ and the vector $\Theta_*$. Such a notation will be frequently used in the following derivations. As a result, let us define $\Delta \overline{\theta}_{t+1}\defn \sum_{m=1}^{N}\frac{ |\mathcal{S}_{m,t}|}{|\mathcal{S}_t|}\Delta\theta_{m,t+1}$ and thereby we can rewrite~\eqref{Eqn:GlobalLearnModel} as
\begin{align*}
	\Delta\Theta_{t+1} &= (1-\beta_t) \Delta\Theta_t + \beta_t \Delta \overline{\theta}_{t+1}.
\end{align*}
This gives rise to the following inequality
\begin{align*}
\|\Delta\Theta_{t+1}\|&\leq  (1-\beta_t) \|\Delta\Theta_t\| + \beta_t \|\Delta \overline{\theta}_{t+1}\|
\end{align*}
due to the triangle inequality, where $\|\mathbf{z}\|$ denotes the (any) norm of vector $\mathbf{z}$. Thus, we can further obtain
\begin{align*}
	&\mathbb{E}\left[\|\Delta\Theta_{t+1}\|^2\right] \leq (1-\beta_t)^2 \mathbb{E}\left[\|\Delta\Theta_t\|^2 \right]+2\beta_t(1-\beta_t) \\
	&\times \mathbb{E}\left[\|\Delta\Theta_t\|  \|\Delta \overline{\theta}_{t+1}\|\right]+\beta^2_t \mathbb{E}\left[\|\Delta \overline{\theta}_{t+1}\|^2\right].
\end{align*}
This means that $\mathbb{E}\left[\|\Delta\Theta_t\|^2\right]$ eventually converges to zero as long as we ensure $\lim_{t\rightarrow\infty}\mathbb{E}\left[\|\Delta \overline{\theta}_t\|^2\right]=0$ and $\beta_t<1$ in that $\lim_{t\rightarrow\infty}\mathbb{E}\left[\|\Delta \overline{\theta}_t\|^2\right]=0$ makes $\|\Delta\bar{\theta}_t\|$ converge to zero and $\beta_t<1$ makes $\|\Delta \Theta_t\|$ reduce to zero almost surely as time goes to infinity. In order to ensure $\mathbb{E}\left[\|\Delta \overline{\theta}_t\|^2\right]$ eventually converge to zero, we need to have $\lim_{t\rightarrow\infty}\mathbb{E}[\|\Delta \theta_{m,t}\|^2]=\lim_{t\rightarrow\infty}\epsilon_{m,t}=0$ for all $m$ because it guarantees $\|\Delta\theta_{m,t}\|$ reduce to zero, which leads to $\lim_{t\rightarrow\infty}\mathbb{E}\left[\|\Delta\bar{\theta}_t\|^2\right]=0$ when considering $\mathbb{E}\left[\|\Delta\bar{\theta}_t\|^2\right]\leq (\sum_{m=1}^{N}\frac{|\mathcal{S}_{m,t}|}{|\mathcal{S}_{t}|}\|\Delta\theta_{m,t}\|)^2$.

Next, $\Delta\theta_{m,t}$ in~\eqref{Eqn:LocalLearnModel} can be explicitly written as
\begin{align*}
	\Delta\theta_{m,t+1} &=\Delta\Theta_{m,t}-\alpha_m\mathbf{g}_m\left(\Delta\Theta_{m,t}+\Theta_*\right)\\
	&\stackrel{(a)}{\approx}\Delta\Theta_{m,t}-\alpha_m \left[\mathbf{g}_m(\Theta_*)+\nabla\mathbf{g}_m(\Theta_*)\Delta\Theta_{m,t} \right]\\
	&\stackrel{(b)}{=}\left[\mathbf{I}_d-\alpha_m\nabla\mathbf{g}_m(\Theta_*)\right]\Delta\Theta_{m,t},
\end{align*}
where $(a)$ is obtained by using the first-order Taylor's expansion of $\mathbf{g}_m(\Delta\Theta_{m,t}+\Theta_*)$ and $(b)$ is due to considering $\mathbf{g}_m(\Theta_*)=\mathbf{0}$ and a $d\times d$ identity matrix $\mathbf{I}_d$. Hence, the covariance matrix of $\Delta\theta_{m,t}$, denoted by $\mathbf{P}_{m,t}\defn\mathbb{E}[\Delta \theta_{m,t}\Delta \theta^T_{m,t}]$, can be expressed as 
\begin{align}
\mathbf{P}_{m,t+1} =\mathbf{Q}_m \mathbb{E}\left[\Delta\Theta_{m,t}\Delta\Theta^T_{m,t}\right]\mathbf{Q}_m^T, \label{Eqn:App:CovMatDelthe}
\end{align}
where $\mathbf{Q}_m\defn \mathbf{I}_d-\alpha_m\nabla\mathbf{g}(\Theta_*)$ and $T$ denotes the transpose notation of vectors and matrices. Moreover, we know
\begin{align*}
\mathbb{E}\left[\Delta\Theta_{m,t}\Delta\Theta^T_{m,t}\right]= &(1-q_m)\mathbb{E}[\Delta\Theta_t\Delta\Theta^T_t]\\
&+q_m\mathbb{E}[\Delta\widehat{\Theta}_{m,t}\Delta\widehat{\Theta}^T_{m,t}]
\end{align*}
because $q_m=\mathbb{P}[\gamma_{m,t}=0]$. In addition, we know $\Delta\widehat{\Theta}_{m,t}=\widehat{\Theta}_{m,t}-\Theta_t+\Theta_t-\Theta_*=\Delta\Theta_t-(\Delta\Theta_t-\Delta\widehat{\Theta}_{m,t})$ whereas $\Delta\Theta_t$ and $(\Delta\Theta_t-\Delta\widehat{\Theta}_{m,t})$ are orthogonal. We thus have $\mathbb{E}[\Delta\widehat{\Theta}_{m,t}\Delta\widehat{\Theta}^T_{m,t}]=\mathbb{E}[\Delta\Theta_t\Delta\Theta^T_t]+\mathbb{E}[(\Delta\Theta_t-\Delta\widehat{\Theta}_{m,t})(\Delta\Theta_t-\Delta\widehat{\Theta}_{m,t})^T]$ so that $\mathbb{E}[\Delta\Theta_{m,t}\Delta\Theta^T_{m,t}]$ can be expressed as
\begin{align*}
&\mathbb{E}[\Delta\Theta_{m,t}\Delta\Theta^T_{m,t}] =\mathbb{E}[\Delta\Theta_t\Delta\Theta^T_t]+q_m\\
&\mathbb{E}[(\Theta_t-\widehat{\Theta}_{m,t})(\Theta_t-\widehat{\Theta}_{m,t})^T].
\end{align*}
Since $\Theta_t-\Theta_*$ is ``less random" than $\theta_{m,t}-\Theta_*$, we know $\mathbb{E}[\Delta\Theta_t\Delta\Theta^T_t]\leq \mathbb{E}[\Delta\theta_{m,t}\Delta\theta^T_{m,t}]=\mathbf{P}_{m,t}$; i.e., $\mathbf{P}_{m,t}-\mathbb{E}[\Delta\Theta_t\Delta\Theta^T_t]$ is positive semi-definite. Now consider the mean square estimation error of $\Theta_t$ at $D_m$, i.e., $\mathbb{E}[\|\Theta_t-\widehat{\Theta}_{m,t}\|^2]$ can be bounded by $\delta_m(\mathbb{E}[\|\Delta\Theta_t\|^2])$ for all $t$ such that $\delta_m\mathbb{E}[\Delta\Theta_t\Delta\Theta^T_t]\geq \mathbb{E}[(\Theta_t-\widehat{\Theta}_{m,t})(\Theta_t-\widehat{\Theta}_{m,t})^T]$. Therefore, we have
\begin{align}
\mathbb{E}[\Delta\Theta_{m,t}\Delta\Theta^T_{m,t}] &\leq \mathbf{P}_{m,t}+q_m\delta_m\mathbb{E}[\Delta\Theta_t\Delta\Theta^T_t]\nonumber\\
&\leq (1+q_m\delta_m)\mathbf{P}_{m,t} \label{Eqn:App:DelTheCovMat}
\end{align}
due to $\mathbb{E}[\Delta\Theta_t\Delta\Theta^T_t]\leq \mathbf{P}_{m,t}$. According to~\eqref{Eqn:App:CovMatDelthe} and~\eqref{Eqn:App:DelTheCovMat}, we can conclude the following:
\begin{align}
\mathbf{P}_{m,t+1} \leq (1+q_m\delta_m)\mathbf{Q}_m\mathbf{P}_{m,t+1}\mathbf{Q}_m, \label{Eqn:App:IneqCovMatPm}
\end{align}
which leads to
\begin{align}
\|\mathbf{P}_{m,t+1}\| \leq \left(\sqrt{(1+q_m\delta_m)}\|\mathbf{Q}_m\|\right)^2\|\mathbf{P}_{m,t}\|,
\end{align}
where $\|\cdot\|$ is any matrix norm. When $\sqrt{(1+q_m\delta_m)}\|\mathbf{Q}_m\|$ is smaller than unity, $\|\mathbf{P}_{m,t}\|$ will converge to zero as $t$ goes to infinity. Namely, $\epsilon_{m,t}\rightarrow0$ as $\sqrt{(1+q_m\delta_m)}\|\mathbf{Q}_m\|<1$ for all $D_m\in\mathcal{D}$. Since all the matrix norms are equivalent, we choose $\|\mathbf{Q}_m\|_2=\sigma_m(\Theta)$, i.e., the spectral norm of  $\mathbf{Q}_m$. As a result, STFL possesses the learning capability as long as the inequality in~\eqref{Eqn:ConvergeCond} holds.

\subsection{Proof of Theorem~\ref{Thm:TimeConstSTFL}}\label{App:ProofAsyConvSTFL}
Suppose the eigenvalue decomposition of $\mathbf{Q}_m$ is $\mathbf{V}_m\mathbf{\Lambda}_m\mathbf{V}^T_m$, where $\mathbf{V}_m$ is a unitary matrix of $\nabla_{\Theta}\mathbf{g}_m(\Theta)$ and $\mathbf{\Lambda}_m=\text{diag}\{(1-\alpha_m\lambda_{m,1}) \cdots (1-\alpha_m\lambda_d)\}$. According to~\eqref{Eqn:App:IneqCovMatPm} and $\text{tr}(\mathbf{P}_{m,t})=\epsilon_{m,t}$,  we can obtain
\begin{align*}
\epsilon_{m,t+1} &\leq (1+q_m\delta_m) \text{tr}\left(\mathbf{V}_m\mathbf{\Lambda}_m\mathbf{V}^T_m\mathbf{P}_{m,t}\mathbf{V}_m\mathbf{\Lambda}_m\mathbf{V}^T_m\right)\\
& \stackrel{(a)}{\leq} (1+q_m\delta_m) \text{tr}\left(\mathbf{\Lambda}^2_m\mathbf{V}_m\mathbf{P}_{m,t}\mathbf{V}^T_m\right) \\
& \stackrel{(b)}{\leq}  (1+q_m\delta_m)\sigma^2_m(\Theta) \text{tr}(\mathbf{P}_{m,t}),
\end{align*} 
where $(a)$ is due to $\text{tr}(\mathbf{AB})=\text{tr}(\mathbf{BA})$ for two matrices $\mathbf{A}$ and $\mathbf{B}$ with the same dimension, and $(b)$ is because $\mathbf{V}_m$ is a unitary matrix such that $\text{tr}(\mathbf{V}_m\mathbf{P}_{m,t}\mathbf{V}^T_m)=\text{tr}(\mathbf{P}_{m,t})$ and $\text{tr}(\mathbf{\Lambda}^2_m\mathbf{P}_{m,t})\leq \sigma^2_m(\Theta)\text{tr}(\mathbf{P}_{m,t})=\sigma^2_m(\Theta)\epsilon_{m,t}$. 
Thus, we obtain
\begin{align*}
	\ln\left(\frac{\epsilon_{m,t}}{\epsilon_{m,t+T}}\right)\geq -T \ln\left( (1+q_m\delta_m)\sigma^2_m(\Theta)\right),
\end{align*}
and then let the above lower bound be greater than or equal to one yields
\begin{align*}
	T \geq \frac{-1}{ \ln(1+q_m\delta_m)+\ln\left(\sigma^2_m(\Theta) \right)}.
\end{align*}
The lower bound on $T$ can be minimized by substituting $\alpha^*_m$ in~\eqref{Eqn:OptLearnRate} into  $\sigma_m(\Theta)$ so as to ensure $\sigma_m(\Theta)$ achieve its maximum value $\sigma^*_m(\Theta)$. Therefore, $\tau_m$ in~\eqref{Eqn:LocalTimeConst} is found according to~\eqref{Eqn:DefnLocalTimeConst2} and $\tau$ in~\eqref{Eqn:LearnTimeConst} is readily obtained based on the definition.


\bibliographystyle{IEEEtran}
\bibliography{IEEEabrv,Ref_STFL} 

\begin{thebibliography}{10}
\providecommand{\url}[1]{#1}
\csname url@samestyle\endcsname
\providecommand{\newblock}{\relax}
\providecommand{\bibinfo}[2]{#2}
\providecommand{\BIBentrySTDinterwordspacing}{\spaceskip=0pt\relax}
\providecommand{\BIBentryALTinterwordstretchfactor}{4}
\providecommand{\BIBentryALTinterwordspacing}{\spaceskip=\fontdimen2\font plus
\BIBentryALTinterwordstretchfactor\fontdimen3\font minus
  \fontdimen4\font\relax}
\providecommand{\BIBforeignlanguage}[2]{{%
\expandafter\ifx\csname l@#1\endcsname\relax
\typeout{** WARNING: IEEEtran.bst: No hyphenation pattern has been}%
\typeout{** loaded for the language `#1'. Using the pattern for}%
\typeout{** the default language instead.}%
\else
\language=\csname l@#1\endcsname
\fi
#2}}
\providecommand{\BIBdecl}{\relax}
\BIBdecl

\bibitem{SCHULZ17}
P.~Schulz, M.~Matthe, H.~Klessig, M.~Simsek, G.~Fettweis, J.~Ansari, S.~A.
  Ashraf, B.~Almeroth, J.~Voigt, I.~Riedel, A.~Puschmann, A.~Mitschele-Thiel,
  M.~Muller, T.~Elste, and M.~Windisch, ``Latency critical {IoT} applications
  in {5G}: Perspective on the design of radio interface and network
  architecture,'' \emph{{IEEE} Commun. Mag.}, vol.~55, no.~2, pp. 70--78, Feb.
  2017.

\bibitem{KC15wc}
S.-Y. Lien, S.-C. Hung, K.-C. Chen, and Y.-C. Liang, ``Ultra-low latency
  ubiquitous connections in heterogeneous cloud radio access networks,''
  \emph{{IEEE} Wireless Commun.}, vol.~22, no.~3, pp. 22--31, Jun. 2015.

\bibitem{YQLHJCXG19}
Y.~Qian, L.~Hu, J.~Chen, X.~Guan, M.~M. Hassan, and A.~Alelaiwi,
  ``Privacy-aware service placement for mobile edge computing via federated
  learning,'' \emph{Information Sciences}, vol. 505, pp. 562--570, 2019.

\bibitem{JRHWTH19}
J.~Ren, H.~Wang, T.~Hou, S.~Zheng, and C.~Tang, ``Federated learning-based
  computation offloading optimization in edge computing supported internet of
  things,'' \emph{IEEE Access}, no.~7, pp. 69\,194--69\,201, 2019.

\bibitem{KYTJYS2020}
K.~Yang, T.~Jiang, Y.~Shi, and Z.~Ding, ``Federated learning via over-the-air
  computation,'' \emph{{IEEE} Trans. Wireless Commun.}, vol.~19, no.~3, pp.
  2022--2035, Mar. 2020.

\bibitem{JRENE2019}
J.~{Ren}, H.~{Wang}, T.~{Hou}, S.~{Zheng}, and C.~{Tang}, ``Federated
  learning-based computation offloading optimization in edge
  computing-supported internet of things,'' \emph{IEEE Access}, vol.~7, pp.
  69\,194--69\,201, 2019.

\bibitem{SWE2019}
S.~{Wang}, T.~{Tuor}, T.~{Salonidis}, K.~K. {Leung}, C.~{Makaya}, T.~{He}, and
  K.~{Chan}, ``Adaptive federated learning in resource constrained edge
  computing systems,'' \emph{{IEEE} J. Sel. Areas Commun.}, vol.~37, no.~6, pp.
  1205--1221, Jun. 2019.

\bibitem{DG2019}
D.~G{\"{u}}nd{\"{u}}z, P.~de~Kerret, N.~D. Sidiropoulos, D.~Gesbert, C.~R.
  Murthy, and M.~van~der Schaar, ``Machine learning in the air,'' \emph{{IEEE}
  J. Sel. Areas Commun.}, vol.~37, no.~10, pp. 2184--2199, Oct. 2019.

\bibitem{MMADG20-1}
M.~M. Amiri and D.~G\"{u}nd\"{u}z, ``Federated learning over wireless fading
  channels,'' \emph{{IEEE} Trans. Wireless Commun.}, vol.~19, no.~5, pp.
  3546--3557, May 2020.

\bibitem{MMADG20-2}
------, ``Machine learning at the wireless edge: Distributed stochastic
  gradient descent over-the-air,'' \emph{{IEEE} Trans. Signal Process.},
  vol.~68, pp. 2155--2166, Mar. 2020.

\bibitem{GZYDDG21}
G.~Zhu, Y.~Du, D.~Gündüz, and K.~Huang, ``One-bit over-the-air aggregation
  for communication-efficient federated edge learning: Design and convergence
  analysis,'' \emph{{IEEE} Trans. Wireless Commun.}, vol.~20, no.~3, pp.
  2120--2135, Mar. 2021.

\bibitem{CLGLPKV21}
C.~Li, G.~Li, and P.~K. Varshney, ``Communication-efficient federated learning
  based on compressed sensing,'' \emph{IEEE Internet Things J.}, vol.~8,
  no.~20, pp. 15\,531--15\,541, Apr. 2021.

\bibitem{HBMEMDR16}
H.~B. McMahan, E.~Moore, D.~Ramage, S.~Hampson, and B.~A. Arcas,
  ``Communication-efficient learning of deep networks from decentralized
  data,'' \emph{arXiv preprint: arXiv:1602.05629}, 2016.

\bibitem{JKHBMFXY16}
J.~Konecný, H.~B. McMahan, F.~X. Yu \emph{et~al.}, ``Federated learning:
  Strategies for improving communication efficiency,'' in \emph{the 29th
  Conference on Neural Information Processing Systems (NIPS)}, 2016, pp. 1--6.

\bibitem{ABJFPY13}
A.~Bensoussan, J.~Frehse, and P.~Yam, \emph{Mean Field Games and Mean Field
  Type Control Theory}, 1st~ed.\hskip 1em plus 0.5em minus 0.4em\relax
  Springer-Verlag, 2013.

\end{thebibliography}

\end{document}